%% file: main.tex
\documentclass[10pt,twocolumn,letterpaper]{article}

\usepackage[pagenumbers]{wacv} 

\input{preamble}

\definecolor{wacvblue}{rgb}{0.21,0.49,0.74}
\usepackage[pagebackref,breaklinks,colorlinks,allcolors=wacvblue]{hyperref}

\def\wacvPaperID{491} 
\def\confName{WACV}
\def\confYear{2026}

\title{ZS-VCOS: Zero-Shot Video Camouflaged Object Segmentation By Optical Flow and Open Vocabulary Object Detection}

\author{
    Wenqi Marshall Guo$^{1,2}$   \quad Mohamed Shehata$^{1}$
     \quad Shan Du$^{1,*}$\\
    $^1$Department of CMPS, University of British Columbia, Canada \\
    $^2$Group of Methane Emission Observation \& Warning (MEOW) , Weathon Software, Canada \\
    {\tt\small wg25r@student.ubc.ca, mohamed.sami.shehata@ubc.ca, shan.du@ubc.ca}
    {\small *Corresponding Author}
}

\begin{document}
\flushbottom
\maketitle
\input{sec/0_abstract}

\input{sec/1_intro}
\input{sec/2_related_work}

\input{sec/3_method}

\input{sec/4_exp}
\input{sec/5_con}
\newpage
{
    \small
    \bibliographystyle{ieeenat_fullname}
    \bibliography{main}
}

\input{sec/X_suppl}

\end{document}

%% file: preamble.tex
\usepackage{amssymb}
\usepackage{array} 
\usepackage{amsmath}
\usepackage{graphicx}
\usepackage{float}
\usepackage[linesnumbered,ruled,vlined,longend]{algorithm2e}
\usepackage{footnote}
\makesavenoteenv{tabular}
\makesavenoteenv{table}
\usepackage{pgfplots}
\pgfplotsset{compat=1.17}
\usepackage{tikz}
\usepackage{bbding}
\usepackage{xcolor}

%% file: sec/0_abstract.tex
\begin{abstract}
Camouflaged object segmentation presents unique challenges compared to traditional segmentation tasks, primarily due to the high similarity in patterns and colors between camouflaged objects and their backgrounds. Effective solutions to this problem have significant implications in critical areas such as pest control, defect detection, and lesion segmentation in medical imaging. Prior research has predominantly emphasized supervised or unsupervised pre-training methods, leaving zero-shot approaches significantly underdeveloped. Existing zero-shot techniques commonly utilize the Segment Anything Model (SAM) in automatic mode or rely on vision-language models to generate cues for segmentation; however, their performances remain unsatisfactory, due to the similarity of the camouflaged object and the background. This work studies how to avoid training by integrating large pre-trained models like SAM-2 and Owl-v2 with temporal information into a modular pipeline. Evaluated on the MoCA-Mask dataset, our approach achieves outstanding performance improvements, significantly outperforming existing zero-shot methods by raising the F-measure ($F_\beta^w$) from 0.296 to 0.628. Our approach also surpasses supervised methods, increasing the F-measure from 0.476 to 0.628. Additionally, evaluation on the MoCA-Filter dataset demonstrates an increase in the success rate from 0.628 to 0.697 when compared with FlowSAM, a supervised transfer method. A thorough ablation study further validates the individual contributions of each component. Besides our main contributions, we also highlight inconsistencies in previous work regarding metrics and settings. \footnote{Code can be found on GitHub after publication.}
\end{abstract}

%% file: sec/1_intro.tex
\section{Introduction}
\label{sec:intro}
\begin{figure}
    \centering
    \includegraphics[width=0.9\linewidth]{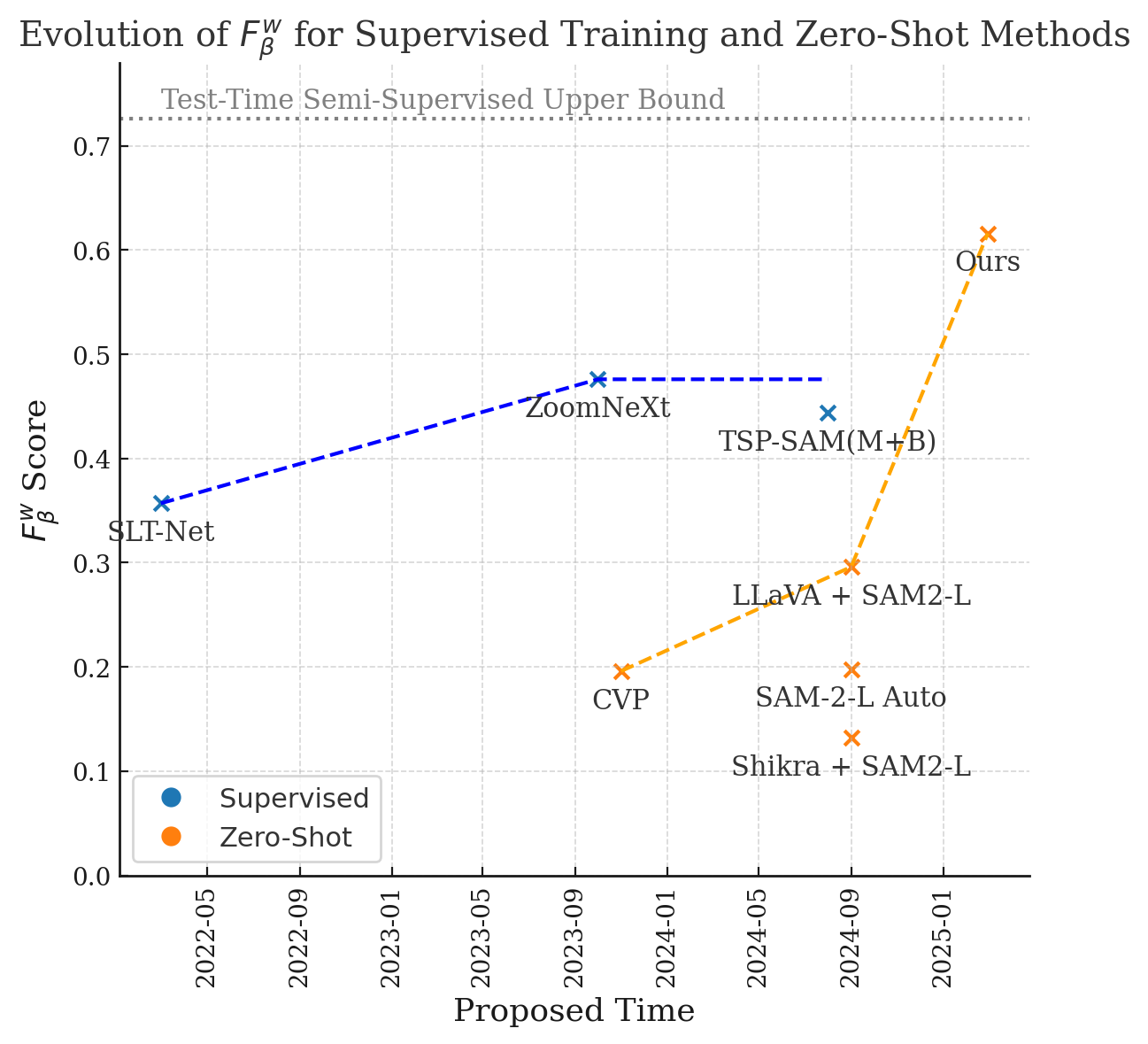}
    \caption{\textbf{Evolution of \( F_{\beta}^{w} \) scores over time for supervised and zero-shot methods on an animal dataset.} The \( F_{\beta}^{w} \) metric is selected for its representativeness and consistent use across all comparison methods. Most zero-shot approaches utilize prior knowledge by explicitly instructing models to detect animals, except for CVP and SAM-2L Auto. Our zero-shot method notably surpasses all previous zero-shot approaches and even outperforms supervised methods, achieving performance close to the test-time semi-supervised upper bound.}
    \label{fig:enter-label}
\end{figure}
\begin{figure*}
    \centering
    \includegraphics[width=1\linewidth]{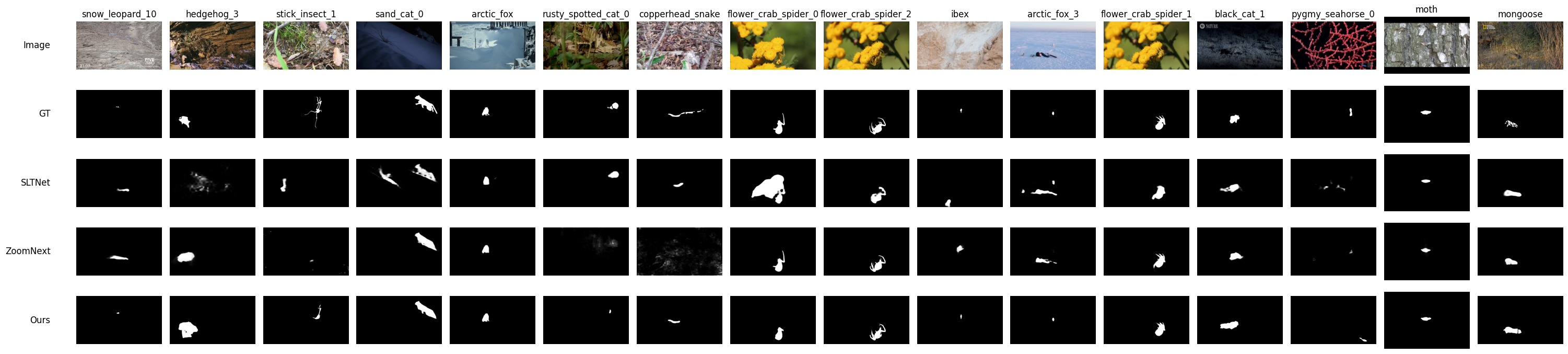}
    \caption{Visual Comparison Of Our Methods and Previous Supervised Methods.}
    \label{fig:examples}
\end{figure*}

Camouflaged object detection and segmentation (COD and COS) is an image detection/segmentation task for objects that are concealed in the background (See Figure \ref{fig:examples} for example). It poses significant challenges beyond traditional object detection and segmentation tasks. This increased difficulty primarily stems from the inherent nature of camouflaged objects, which are visually similar to their backgrounds in terms of patterns, colors, and textures \cite{sltnet,xiao_survey_2024}, effectively blending into their surroundings and complicating accurate identification and delineation. Despite these challenges, effective solutions for COD and COS have considerable real-world significance, especially in critical fields such as defect detection \cite{Kumar_2008}, pest control \cite{Rustia_Lin_Chung_Zhuang_Hsu_Lin_2020}, and medical imaging for lesion segmentation \cite{Fan_Ji_Zhou_Chen_Fu_Shen_Shao_2020}.

Extending these image-based tasks into the temporal domain, video camouflage object detection and segmentation (VCOD and VCOS) have emerged as specialized subsets derived from video object detection (VOD) and video object segmentation (VOS), respectively. By leveraging motion cues, such methods can potentially overcome some limitations inherent to static images. Optical flow, for instance, has proven particularly useful by measuring pixel-level movements, thus enabling differentiation of moving camouflaged entities from their backgrounds.

However, camouflage-related tasks in both static and dynamic contexts remain relatively underdeveloped compared to traditional detection and segmentation methods. Most prior work in this area has focused on supervised learning, relying on complex architectures and labelled data. Yet, even these supervised models often struggle with camouflaged objects due to the lack of distinct features. On the other hand, zero-shot methods, which avoid training by using large pre-trained models like SAM and vision-language models, are severely less explored and currently perform worse than supervised methods.

To address this gap, we propose a method that integrates optical flow, a vision-language model, and SAM in a modular pipeline. Each stage of the pipeline uses the output of the previous one to refine its segmentation cues. Rather than relying on any training or fine-tuning, our approach operates in a zero-shot setting and achieves strong performance. On the MoCA-Mask dataset, our method improves mIoU from 0.273 (baseline zero-shot methods) to 0.561. It also outperforms multiple supervised methods, which typically score around 0.422. Furthermore, on the MoCA-Filtered dataset, our method raises the detection success rate from 0.628 to 0.697. These gains highlight the effectiveness of combining motion-based cues with strong foundation models.

In summary, our technical contributions are (1) a zero-shot framework for camouflaged object segmentation in video that surpasses supervised baselines, (2) extensive experimentation on different components and prompting strategies of our methods, and (3) insights demonstrating that properly designed zero-shot pipelines can not only compete with but in some cases outperform traditional supervised approaches. 

Additionally, we noticed that previous works often failed to systematically compare their results against other methods evaluated under the same settings (test-time supervised, also known as tracking, and test-time unsupervised). Furthermore, metric calculation in these benchmarks frequently suffered from inconsistent aggregation methods and inadequate handling of special cases. In this work, we highlight these issues, re-evaluate the state-of-the-art methods using a consistent and corrected metric, and ensure a direct and fair comparison between our method and the current state-of-the-art. We urge the research community to adopt standardized evaluation practices to enable clearer and more meaningful comparisons in future studies.

\begin{figure*}
    \centering
    \includegraphics[width=0.7\linewidth]{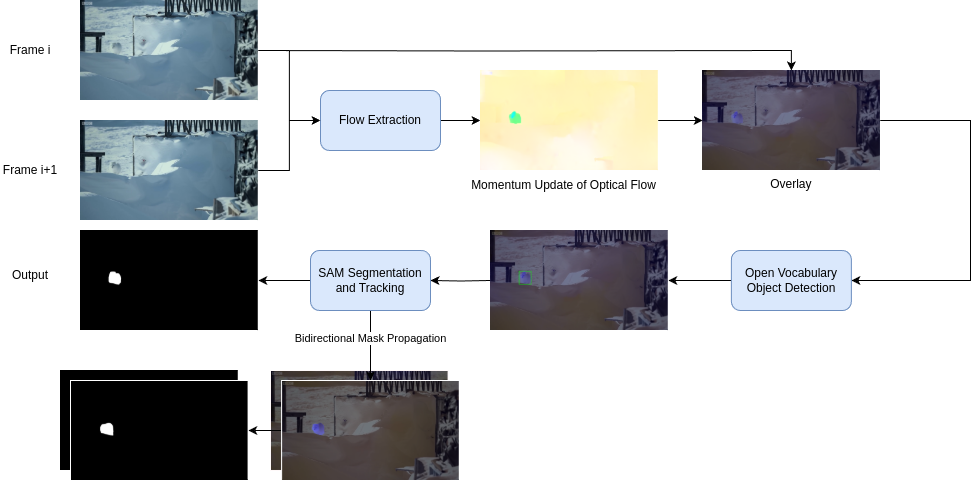}
    \caption{Overview of Our Method.}
    \label{fig:enter-label}
\end{figure*}

%% file: sec/2_related_work.tex
\section{Related Work}
\subsection{Optical Flow} 
Optical flow is a technique used to measure pixel movement in videos. It has been used in video processing or recognition for a long time; one of the most significant works is the two-stream network published in 2014 \cite{Simonyan_Zisserman_2014}. There are two types of optical flow: sparse and dense optical flow. Sparse optical flow gives a movement vector for points of interest in the image, whereas dense optical flow estimates movement for all pixels in the image. One of the most famous sparse optical flows is the Lucas–Kanade method \cite{10.5555/1623264.1623280}, which uses the assumption that local pixels have similar motion. It can be used in camera motion estimation for panoramic image generation or motion compensation. Dense optical methods, like RAFT \cite{raft} and GMFlow \cite{Xu_Zhang_Cai_Rezatofighi_Tao_2022}, can provide movement information for every pixel in the frame, and it has demonstrated promising performance in camouflaged object detection based on movement differences between foreground and background, although their methods relay on training and the perfomance can be furture improved. 
\newcolumntype{G}{>{\color{gray}}c}
\begin{table*}
    \centering
    \begin{tabular}{c|c|c|ccc|GGG}
    \hline
    Method&  Pub.&Setting& $S_{\alpha}\uparrow$& $F_{\beta}^{w}\uparrow$& MAE$\downarrow$ & $E_{\phi} \uparrow$& mDice$\uparrow$&mIoU$\uparrow$\\
    \hline
 SLT-Net \cite{sltnet} &  CVPR 22&SV Tr& 0.656& 0.357& 0.021 & 0.785& 0.397&0.310\\
 ZoomNeXt \cite{pang_zoomnext:_2024} &  TPAMI 24&SV Tr& \textbf{0.734}& \textbf{0.476}& 0.010 & 0.736&\textbf{ 0.497}&\textbf{0.422}\\
 TSP-SAM(M+B) \cite{sam_eyes}&  CVPR 24&SV Tr&0.689& 0.444 & \textbf{0.008} & \textbf{0.808}& 0.458&0.388\\
 Gao \textit{et. al} \cite{Gao_Feng_Wang_Hong_Zhou_Fei_Wang_Zhang_2025}& arXiv 25& SV Tr& 0.706& 0.455& 0.011& -& 0.495& 0.404\\
 \hline
  SAM2 Tracking \cite{sam2_vcod}&  arXiv 24&SV Te& 0.804& 0.691& 0.004 & -& -&-\\
  \hline
   SAM-PM \cite{Meeran_T_Mantha_2024} & CVPRW 24& SV Tr+Te& 0.728& 0.567&  0.009 & 0.813& 0.594&0.502\\
   Finetuned SAM2-T + Prompts \cite{Zhou_Sun_Li_Benini_Konukoglu_2024}&  arXiv 24&SV Tr+Te& \textbf{0.832}& \textbf{0.726}& \textbf{0.005} & \textbf{0.908}&  \textbf{0.756}&\textbf{0.652}\\
 \hline
 CVP \cite{llm_vcod}&  ACM MM 24&ZS& 0.569& 0.196& 0.031 & -& -&-\\
 SAM-2-S Auto \cite{Zhou_Sun_Li_Benini_Konukoglu_2024}&  arXiv 24&ZS& 0.497& 0.201& 0.141& 0.608& 0.202&0.174\\
 LLaVA + SAM2-L \cite{Zhou_Sun_Li_Benini_Konukoglu_2024}&  arXiv 24&ZS w/ PK& 0.624& 0.315&  0.046& 0.688& 0.334&0.291\\
 Shikra + SAM2-L \cite{Zhou_Sun_Li_Benini_Konukoglu_2024}&  arXiv 24&ZS w/ PK& 0.502& 0.146& 0.107 & 0.590& 0.157&0.124\\
 Ours&  -&ZS w/ PK& \textbf{0.776}& \textbf{0.628}& \textbf{0.008} & \textbf{0.878}& \textbf{0.648}&\textbf{0.550}\\
    
    \end{tabular}
    \caption{\textbf{Performance comparison on the MoCA-Mask dataset \cite{sltnet}.} ``SV Tr" denotes supervised training. ``SV Te" denotes supervised testing, where one frame from the video was provided to the model along with prompts. ``ZS" indicates zero-shot learning, while ``ZS w/ PK" means zero-shot with prior knowledge (since the model already knows it is looking for animals). The grouping of methods is based on settings. Metrics shown in gray represent results from prior work that may contain methodological inconsistencies. These are included for transparency and completeness but should be interpreted with caution. Our method significantly outperforms all zero-shot and even supervisely trained and unsupervisely tested methods.}
    \label{tab:res}
\end{table*}

\subsection{Moving Object Segmentation} 
Moving object segmentation is a task aiming to segment moving objects within a video sequence. These objects could be general entities, as in DAVIS \cite{davis} and YouTube-VOS \cite{youtubevos}, or camouflaged ones, as presented in MoCA-Mask \cite{sltnet} and CAD \cite{cad}. The segmentation task can be performed in two scenarios: test-time semi-supervised, where one annotated frame is provided and the model propagates this annotation to subsequent frames, and test-time unsupervised, where no annotation is provided during testing. These two methods have distinct difficulty levels and should be compared separately. 

Optical flow has been extensively used in video object segmentation, primarily in two ways: propagating segmentation masks and differentiating objects from background based on different motions.

\subsubsection{Methods Leveraging Optical Flow as Motion Cues} 

Brox \textit{et al.} \cite{Brox_Malik_2010} utilized long-term optical flow trajectories combined with clustering for segmenting videos. Ochs \textit{et al.} \cite{Ochs_Malik_Brox_2014} applied flow-based motion cues to resolve ambiguities in color-based segmentation. Xiao \textit{et al.} \cite{Xiao_2018_CVPR} employed optical flow cues to reinforce target frame representations. Yang \textit{et al.} \cite{optical_seg} used both optical flow and RGB input to assist video object segmentation. FlowI-SAM and FlowP-SAM \cite{flowsam} utilized optical flow either exclusively as input (FlowI-SAM) or as a prompt guiding segmentation of RGB frames (FlowP-SAM). Both FlowI-SAM and FlowP-SAM handled standard and camouflaged objects effectively.

\subsubsection{Methods Employing Optical Flow for Mask Propagation}
Tsai \textit{et al.} \cite{Tsai_2016_CVPR} considered segmentation and optical flow simultaneously, using optical flow to propagate masks and segmentation masks to refine flow boundaries. TR-OVIS \cite{Yan_Sundermeyer_Tan_Lu_Tombari_2024} employed optical flow to propagate key-frame information, thus enhancing inference speed for open-vocabulary video instance segmentation.

\subsubsection{Joint Modeling of Segmentation and Flow without Using Flow as Input}
Cheng \textit{et al.} \cite{Cheng_2017_ICCV} (SegFlow) treated segmentation and optical flow estimation as similar tasks and jointly trained a network to take in video frames and output segmentation masks and optical flow.

\subsubsection{Alternative Motion Methods without Optical Flow}

LangGas \cite{guo_langgas:_2025} applied background subtraction to isolate moving regions, followed by an open vocabulary object detector and SAM2 \cite{sam2} to segment gas leaks in synthetic datasets. Zero-shot Background Subtraction (ZBS) \cite{An_Zhao_Yu_Guo_Zhao_Tang_Wang_2023} detected the displacement of objects across frames using object detection techniques to classify their motion status, thus identifying moving objects without optical flow.

Wang \textit{et al.} \cite{Wang_2023_ICCV}, Wang \textit{et al.} \cite{Wang_Yan_Chen_Jiang_Tang_Hu_Kang_Xie_Gavves_2024}, and Li \textit{et al.} \cite{Li_Liu_Sun_Wu_Zhang_Zhu_2024} performed segmentation directly from raw RGB frames with text as queries, without incorporating explicit motion signals or optical flow.

\subsection{Moving Camera Background Subtraction} 
Moving camera background subtraction (MCBS) is very similar to the VOS task, where it extracts the moving foreground from the background by using a background model. Unlike fixed camera background subtraction, where pixels from the same object/background are mostly aligned throughout the video, MCBS is challenging as the background is moving, and the algorithm cannot simply compare the pixel value at the same absolute location. Kurnianggoro \textit{et al.} \cite{Kurnianggoro_Wahyono_Yu_Hernandez_Jo_2016} used motion compensation to solve this problem, while DeepMCBM \cite{Erez_Weber_Freifeld_2022} and PanoramicPCA \cite{Moore_Gao_Nadakuditi_2019} builts a panoramic background model. 

\subsection{Camouflage Object Detection and Segmentation} 
Unlike regular object detection and segmentation, camouflage object tasks are significantly more challenging because the foreground usually seamlessly blends into the background. There are two tasks in camouflage object detection: image-based camouflage object detection (usually referred to as COD) and video-based camouflage object detection (VCOD). They could also be extended to segmentation, namely COS and VCOS. VCOD/S allows the model to use motion cues to detect the foreground but also brings in the challenges of temporal changes \cite{xiao_survey_2024}.

\subsubsection{Datasets}
Since this paper focuses on VCOS, we mainly introduce video-based datasets here. For image-based datasets like COD10K \cite{COD10K}, N4K \cite{NC4K}, and CAMO \cite{CAMO}, readers can refer to the review article \cite{xiao_survey_2024}.

There are two major datasets and 3 variants in video camouflage object detection: Camouflaged Animal Dataset (CAD) \cite{cad} and Moving Camouflaged Animal Dataset (MoCA) \cite{Lamdouar_Yang_Xie_Zisserman_2020}. However, MoCA is an object detection dataset but not a segmentation dataset. It contains some non-camouflaged animals or animals that do not have locomotion. Thus, two variances of MoCA were proposed. MoCA-filtered \cite{Yang_Lamdouar_Lu_Zisserman_Xie_2021} and MoCA-Mask \cite{sltnet}.

MoCA-filtered mainly removed non-locomotive videos from the dataset, with additional processing such as cropping away logos and borders, resampling frames, and incorporating the bounding boxes. Since it still lacks segmentation masks, papers using this dataset  (\cite{Yang_Lamdouar_Lu_Zisserman_Xie_2021} \cite{flowsam}) used the detection success rate based on the IoU threshold to evaluate the results. MoCA-Mask improved MoCA by removing scenes with obvious animals and converting bounding boxes into masks. In addition to ground truth masks provided every 5 frames, they also used bidirectional optical flow to generate pseudo masks for unlabelled frames.

\subsubsection{Algorithms} 
Existing methods can be classified into supervised, unsupervised, and zero-shot categories based on training settings and into test-time semi-supervised or test-time unsupervised categories based on inference settings.

SLT-Net \cite{sltnet} is a supervisely trained and unsupervisely tested model. It argued that when using optical flow and homography, the error might be accumulated from both the motion estimation and segmentation. Thus, they proposed to use a unified framework for both motion estimation and segmentation. Additionally, they used a long-term spatiotemporal transformer to refine short-term predictions, although this long-term module provides marginal improvement. ZoomNeXt \cite{pang_zoomnext:_2024} is an improved version of ZoomNet \cite{Pang_Zhao_Xiang_Zhang_Lu_2022}, mainly adapted from image-based COS to video-based COS and improved performance by introducing more structural extensions. ZoomNeXt is trained on both image COS datasets and video COS datasets, including MoCA-Mask \cite{sltnet}. ZoomNet and ZoomNeXt both use zooming to capture features at different scales. Similar to SLT-Net, they are both supervisely trained and unsupervisely tested methods.

Previous studies have sometimes failed to clearly differentiate between test-time semi-supervised and unsupervised tasks, despite their differing levels of difficulty. For example, SAM-PM \cite{Meeran_T_Mantha_2024}, requiring supervision during both training and inference, reported state-of-the-art results compared with SLT-Net \cite{sltnet}. However, SLT-Net operates under supervised training but unsupervised testing conditions. This fundamental difference in evaluation criteria renders direct comparisons between these two methods somewhat inequitable. Although the authors of SAM-PM described their method as semi-supervised (which we refer to in this paper as test-time supervised), they did not clearly acknowledge this distinction when making comparisons or drawing conclusions. 

Flow-SAM \cite{flowsam} and Motion Grouping \cite{Yang_Lamdouar_Lu_Zisserman_Xie_2021}, though trained initially for video object segmentation tasks, demonstrated robust performance on VCOS tasks. Specifically, Flow-SAM utilizes supervised training, while Motion Grouping employs self-supervised training. Neither method requires supervision during inference.

For zero-shot unsupervised testing neither training nor inference is supervised), Chain of Vision Perception (CVP) \cite{llm_vcod} represents an early effort employing vision-language models (VLMs) for COD/S tasks, with a primary focus on images rather than videos. CVP prompts a vision-language model to identify the location of camouflaged objects. Subsequently, these locations are refined and given to a segmentation model. Properly designed prompting can further enhance the model's performance. CVP achieved higher performance than several supervised methods on datasets such as CAMO \cite{CAMO}, COD10K \cite{COD10K}, and NC4K \cite{NC4K}. However, its results on the MoCA-Mask were suboptimal, with a weighted F-score ($F_{\beta}^{w}$) of 0.196. Zhou \textit{et al.} \cite{Zhou_Sun_Li_Benini_Konukoglu_2024} improved upon this by employing LLaVA \cite{Liu_Li_Wu_Lee_2023} or Shikra \cite{Chen_Zhang_Zeng_Zhang_Zhu_Zhao_2023} as the vision-language model and utilizing SAM-2 for segmentation. A similar approach is evident in Grounded SAM \cite{Liu_Zeng_Ren_Li_Zhang_Yang_Jiang_Li_Yang_Su_etal._2024}, which integrates Grounding DINO and SAM for open-vocabulary segmentation of regular objects.

While comparing test-time unsupervised methods with semi-supervised methods is inherently unfair, semi-supervised inference methods without prior training have demonstrated that SAM-2 can reasonably track camouflaged objects when provided with accurate prompts.

Detailed performance comparisons of these methods can be found in Table~\ref{tab:res}.


%% file: sec/3_method.tex
\begin{table*}
    \centering
    \begin{tabular}{>{\centering\arraybackslash}p{5mm}|cccc|ccc}
    \hline
         &    Motion Detection & Mean Subtraction&Momentum & Tracking &$S_{\alpha}\uparrow$&  $E_{\phi}\uparrow$& mIoU $\uparrow$\\
                  \hline
 (a)& None& -& -& None & 0.621& 0.596&0.252\\ 
         (b)&    None & -& -& Bidirectional &0.643&  0.657& 0.301\\
         \hline 
         (c)&    OF Only & \Checkmark& & Bidirectional &0.752&  0.832& 0.508\\
 (d)& OF Only & \Checkmark& \Checkmark& Bidirectional & 0.750& 0.824&0.513\\
 \hline
 (e)& OF/BGS& \Checkmark&  & Bidirectional & 0.759& 0.843&0.522\\
 (f)& OF/BGS& \Checkmark& & None & 0.676& 0.698&0.363\\
 (g)& OF/BGS& \Checkmark& \Checkmark& None & 0.683& 0.723&0.372\\

 (h)& OF/BGS& \Checkmark& \Checkmark& Forward Only & 0.747& 0.825&0.497\\
  \hline
 (i)& OF/BGS& & \Checkmark & Bidirectional & \textbf{0.782}& 0.859&\textbf{0.561}\\
  Ours& OF/BGS& \Checkmark&\Checkmark & Bidirectional & 0.776& \textbf{0.878}&0.550\\
    \end{tabular}
  \caption{Ablation study of different components including motion detection (optical flow and background subtraction), mean subtraction, momentum update, and tracking strategies. Motion detection includes either optical flow only (OF) or a combination of optical flow and background subtraction (OF/BGS). We evaluate each configuration using $S_\alpha$, $E_\phi$, and mean IoU (mIoU).}
    \label{tab:ab}
\end{table*}

\section{Proposed Methods}
\subsection{Motion Detection}
Our method builds upon LangGas \cite{guo_langgas:_2025}. Gas leakage shares many similarities with a camouflage object: they both have low contrast against the background, but they often have different relative motion with respect to the background. Previous studies, including LangGas \cite{guo_langgas:_2025} and VideoGasNet \cite{WANG2022121516}, have shown that background subtraction effectively captures subtle changes in the input. High-quality masks can then be extracted from the resulting foreground using vision–language models (VLMs) together with SAM2 \cite{sam2} \cite{guo_langgas:_2025}. However, traditional BGS methods can only be used in fixed camera settings, and most camouflage object segmentation datasets and real-world applications do not feature a fixed camera; while a moving camera background subtraction method can sometimes work, it may fail under complex camera motion. In addition, if an object does not fully move away to expose the background behind it, a valid background model cannot be built. \footnote{Although the object's edges could be shown in the foreground map, camera movements may also highlight these edges, making it hard for the algorithm to distinguish them. }

To address such challenges, we turn to another commonly used motion detection method: optical flow. By tracking the movement of each pixel between two adjacent frames, optical flow can show different movement patterns in the image. Following Motion Grouping \cite{Yang_Lamdouar_Lu_Zisserman_Xie_2021} and FlowSAM \cite{flowsam}, we employ RAFT \cite{raft} to compute optical flow. However, we found that highly repetitive backgrounds or videos with margins can compromise RAFT optical flow, thereby diminishing its usefulness. Thus, we combine optical flow with background subtraction, applying the latter (BGS) when there is no camera motion and using optical flow otherwise. To detect camera motion, we use a simple Lucas–Kanade method \cite{10.5555/1623264.1623280} to track points in the video. The movement is used to estimate the affine transformation throughout the video and detect the furthest point the camera reached. 

For videos processed using optical flow, we compute an optical flow tensor \( F \in \mathbb{R}^{(t-1) \times h \times w \times 2} \), where \( t \) is the total number of frames, \( h \) and \( w \) denote frame height and width, respectively, and the two channels represent horizontal and vertical pixel displacements. The corresponding intensity map is obtained by calculating the magnitude of displacement vectors at each pixel location, and the intensity map is normalized to 0-255, as shown in Equation \eqref{eq:flow_intensity}. We also experimented with maintaining a momentum-based moving average over the flow vector map to address cases where the object temporarily stops moving. The formulation is given in Equation~\eqref{eq:momentum}. We also experimented with subtracting the mean displacement vector (averaged over the frame) from every pixel to reduce camera motion, inspired by the Two-Stream Network approach \cite{Simonyan_Zisserman_2014}.

\begin{equation}
    I_{i,x,y} = \operatorname{normalize}_{[0,255]}(\|F_{i, x, y, :}\|_2)
    \label{eq:flow_intensity}
\end{equation}

\begin{equation}
    F_{i} = \begin{cases}
			F_{i}, & i=1\\
            (1-m) \cdot F_{i} + m \cdot F_{i-1}, & i > 1
		 \end{cases}
    \label{eq:momentum}
\end{equation}

For videos analyzed using background subtraction, we followed \cite{guo_langgas:_2025}. First, we obtain a background model tensor \( B \in \mathbb{R}^{t \times h \times w \times 3} \) using MOG2 \cite{mog2_2, knn_mog2_1}. Here, each frame in the background model matches the dimensions and RGB channels of the input frames. The intensity map in this scenario is computed by taking the absolute pixel-wise difference between the current frame \( C_t \) and the background frame \( B_t \), and normalized to 0-255, as detailed in Equation \eqref{eq:bgs}.

\begin{equation}
    I_{i,x,y} = \operatorname{normalize}_{[0,255]}(\|C_{i,x,y} - B_{i,x,y}\|_1)
    \label{eq:bgs}
\end{equation}

The intensity map is then blended into the current frame using a specific color (e.g. blue) to highlight the moving parts in the current frame.

\subsection{Open Vocabulary Detection}
We used Owlv2 \cite{Minderer_Gritsenko_Houlsby_2024} as our detection vision language model (VLM), same as in LangGas \cite{guo_langgas:_2025}. Since all videos in MoCA are about animals or insects, following \cite{Zhou_Sun_Li_Benini_Konukoglu_2024}, we included that in the prompt. Following LangGas \cite{guo_langgas:_2025}, we used one positive prompt and 3 negative prompts so that when the object is closer to the negative prompts, it can be correctly classified into the negative prompt and reduce interference. We used ``an animal or insect being highlighted in blue" as a positive prompt and ``background", ``logo or sign," and ``plant" as negative prompts. Since the camouflage object segmentation is usually a single object problem, we select the box with the highest score after the VLM. 

\subsection{Segmentation and Tracking}
Given that camouflage can significantly reduce object detection performance for VLM, many frames might result in missed detections. However, previous research \cite{Zhou_Sun_Li_Benini_Konukoglu_2024} \cite{Meeran_T_Mantha_2024} \cite{sam2_vcod} has demonstrated that vanilla SAM-2 \cite{sam2_vcod}, when guided by explicit prompts, can achieve effective object tracking. We utilized this tracking capability by supplying SAM-2 with all prompts obtained from VLM detections and allowed it to propagate these prompts across all video frames. These prompts consist of the bounding boxes generated by the VLM and the center of mass of the intensity map within each bounding box as a point prompt.

Since forward propagation alone limits object tracking to frames following the initial successful detection, we implemented a bidirectional propagation approach. We provided prompts for both the original forward-playing video and its reversed sequence. Masks generated from both directions are combined using an OR operation, producing the final robust masks across the entire video sequence.

%% file: sec/4_exp.tex
\section{Experiments and Results}

\subsection{Benchmark}
\subsubsection{Metrics and Datasets}
In this paper, we examine two variants of the MoCA dataset \cite{Lamdouar_Yang_Xie_Zisserman_2020}: MoCA-Mask \cite{sltnet} and MoCA-Filtered \cite{Yang_Lamdouar_Lu_Zisserman_Xie_2021}. The Camouflage Animal Dataset (CAD) \cite{cad} is not used due to its inaccessibility, as the server is offline and the dataset is not provided by a third party. MoCA-Mask is a segmentation dataset, and our evaluation approach aligns with SLT-Net \cite{sltnet}. We report the following metrics: S-measure ($S_a$) \cite{Cheng_Fan_2021}, weighted F-measure ($F_{\beta}^w$) \cite{Margolin_Zelnik-Manor_Tal_2014}, and Mean Absolute Error (MAE). More details on these metrics can be found in the SLT-Net paper and their original sources. We did not focus on E-measure \cite{Fan_Gong_Cao_Ren_Cheng_Borji_2018}, mean Dice coefficient, and mean Intersection-over-Union (IoU) in our comparison with previous methods because they will yield different results depending on metric calculation methods, which we will explain in the supplementary material Section \ref{sec:metric}. All standard metrics are provided in Table~\ref{tab:res}, with metrics that could be miscalculated by the previous method colored in gray. For our internal comparisons in the ablation study, we primarily focus on a subset of these metrics (using a subset of internal comparisons is used in SLT-Net): S-measure, E-measure, and mean IoU. To ensure consistency and fairness, we adopted the evaluation code and methodology from SLT-Net.

Although there are forum discussions about this issue \cite{ignatius_2018}, there are only a few publications mentioned about this issue \cite{Triki_Torr_Tuia_Tuytelaars_Gool_Yu_Blaschko_2023, guo_langgas:_2025}. We encourage future research to clearly specify their metric calculation methodology and consider adopting this standardized frame-then-video averaging approach to facilitate fair comparisons.

Since the SLT-Net method produces soft outputs with continuous pixel values, they considered multiple thresholds and report a max and mean metric. However, as our method produces binary outputs, we used a single threshold value of 0.5.

MoCA-Filtered \cite{Yang_Lamdouar_Lu_Zisserman_Xie_2021} is a detection dataset. Following \cite{Yang_Lamdouar_Lu_Zisserman_Xie_2021} and \cite{flowsam}, we used the detection success rate based on IoU. Similar to MoCA-Mask, we employed the original evaluation code provided by \cite{Yang_Lamdouar_Lu_Zisserman_Xie_2021}.

\subsubsection{Baselines}
For MoCA-Mask, we selected SLT-Net \cite{sltnet}, ZoomNeXt \cite{pang_zoomnext:_2024}, TSP-SAN (M+B) \cite{sam_eyes}, and the proposed method from the MSVOCD dataset paper \cite{Gao_Feng_Wang_Hong_Zhou_Fei_Wang_Zhang_2025} as our supervised training baselines. For zero-shot baselines, we employed Chain of Visual Perception (CVP) \cite{llm_vcod}, SAM-2-L Auto, and SAM-2 combined with either LLaVA \cite{Liu_Li_Wu_Lee_2023} or Shikra \cite{Chen_Zhang_Zeng_Zhang_Zhu_Zhao_2023}, following the approach described in \cite{Zhou_Sun_Li_Benini_Konukoglu_2024}. Additionally, we utilized three test-time supervised methods \cite{Zhou_Sun_Li_Benini_Konukoglu_2024, Meeran_T_Mantha_2024, sam2_vcod} as performance upper bounds.

For MoCA-Filtered, we adopted FlowSAM (including FlowI-SAM and FlowP-SAM) \cite{flowsam} and Motion Grouping \cite{Yang_Lamdouar_Lu_Zisserman_Xie_2021} as baselines for supervised and self-supervised pretraining, respectively, using non-camouflage object datasets and testing on a camouflage dataset.

\subsubsection{Settings}
Since our method is zero-shot without training, we directly evaluated it on the testing set. For MoCA-Mask, to minimize overfitting hyperparameters on the limited test set while ensuring reasonable performance in the real world, we slightly adjusted hyperparameters manually and swept the VLM threshold (following \cite{guo_langgas:_2025}) from 0.03 to 0.13 (inclusive) at a step of 0.02. For MoCA-Filtered, we employed a fixed threshold of 0.12 tuned by hand. Optical flow was computed using RAFT-Things \cite{raft}, employing the implementation provided by Motion Grouping \cite{Yang_Lamdouar_Lu_Zisserman_Xie_2021} with only forward flow and a frame gap of 1. All input images were passed directly to the model processor without resizing or cropping. The momentum parameter ($m$) was set to 0.9. We used \texttt{Owlv2-Base-Patch16-Ensemble} and \texttt{Sam2.1-Hiera-Small} as our VLM and segmentation models.

\begin{table}[]
    \centering
    \begin{tabular}{ccc|c}
    \hline
         Model&  Pub.& Settings& SR\\ \hline
 FlowI-SAM \cite{flowsam}& ACCV 24& SV Transfer&0.628\\
 FlowP-SAM \cite{flowsam}& ACCV 24& SV Transfer&0.645\\
 Motion Grouping \cite{Yang_Lamdouar_Lu_Zisserman_Xie_2021}& ICCV 21& SS Transfer&0.484\\    \hline
      
 ZS-VCOS& -& ZS w/ PK&\textbf{0.697}\\   \hline
    \end{tabular}
    \caption{Success rate (SR) of detection success rate on MoCA-Filtered \cite{Yang_Lamdouar_Lu_Zisserman_Xie_2021}. ``SV" stands for supervised training, ``SV" stands for self-supervised, and ``ZS" stands for zero-shot. Although the previous three methods are trained on VOS datasets, they are not trained on camouflage object datasets. Our method is not trained on any VOS or camouflage datasets and resulted in the highest SR.}
    \label{tab:filtered}
\end{table}

\subsection{Results}
\subsubsection{MoCA-Mask}
The results of our method compared with previous baselines on MoCA-Mask are presented in Table~\ref{tab:res}. Our approach achieves the highest $F_{\beta}^w$ and $S_a$, as well as the lowest MAE among all methods without test-time prompts (unsupervised at test-time). Specifically, we outperform ZoomNeXt, a supervised method published in 2024 and considered state-of-the-art, by +0.152 on $F_{\beta}^w$ and +0.042 on $S_a$. Compared to previous zero-shot methods leveraging prior knowledge, such as LLaVA + SAM2-L \cite{Zhou_Sun_Li_Benini_Konukoglu_2024}, we obtain improvements of +0.332 in $F_{\beta}^w$ and +0.154 in $S_a$. Moreover, our method is only -0.098 behind in $F_{\beta}^w$ and -0.056 in $S_a$ compared to the test-time supervised upper bound reported in \cite{Zhou_Sun_Li_Benini_Konukoglu_2024}. This indicates that our method is very close to this upper bound. Although our improvement in $S_a$ is moderate, we observed that $S_a$ might not be highly discriminative, as even masks with minimal overlap can achieve scores around 0.40.

\subsubsection{MoCA-Filtered}
Our results for MoCA-Filtered are presented in Table~\ref{tab:filtered}. Our method outperforms Flow-SAM \cite{flowsam} and Motion Grouping \cite{Yang_Lamdouar_Lu_Zisserman_Xie_2021}, which are trained on non-camouflage video segmentation datasets using supervised and unsupervised approaches, respectively. We achieve a detection success rate of 0.697, compared to 0.645 for Flow-SAM and 0.484 for Motion Grouping. The improvement here is not as significant as in MoCA-Mask, which may indicate that combining SAM and optical flow, as done in FlowSAM, is already an effective approach.

\subsection{Video-Level Results}
We examined individual results for each video in the test set. Quantitative results for mIoU are presented in Table~\ref{tab:video} in supplemental material, and qualitative results are shown in Figure~\ref{fig:examples}. Both results indicate that our method succeeded in \texttt{stick\_insect\_1} and \texttt{snow\_leopard\_10}, where ZoomNeXt completely failed. Additionally, our approach successfully captured the target in \texttt{ibex}, whereas the other two methods missed it. In \texttt{arctic\_fox\_3}, our method achieved a significantly higher $F_{\beta}^w$ score and effectively avoided stationary objects. Although our method struggled in \texttt{pygmy\_seahorse\_0}, neither ZoomNeXt nor SLT-Net performed well in this case. In other cases where other methods outperformed ours, the margin of improvement was minimal.

\subsection{Ablation Study}
For our ablation study, we designed 9 configurations, as shown in Table~\ref{tab:ab}. Configuration (a) is a minimum baseline with only object detection, used to compare with prior methods such as LLaVA/Shikra + SAM2 \cite{Zhou_Sun_Li_Benini_Konukoglu_2024}. Our result ($S_\alpha = 0.621$) is nearly identical to theirs ($S_\alpha = 0.622$). In (b), we add bidirectional tracking to (a), which slightly improves $S_\alpha$ by +0.022 and mIoU by +0.049. In (c), we add optical flow and mean subtraction on top of (b), leading to a significant improvement: $S_\alpha$ increases by +0.109 and mIoU by +0.207. In (d), we introduce momentum to (c), resulting in a very small drop in $S_\alpha$ (-0.002) but a minor gain in mIoU (+0.005). In (e), we use both optical flow and background subtraction based on camera movements, along with mean subtraction for optical flow, resulted in a slight increase compared to (d), +0.009 in $S_\alpha$ and 0.009 in mIoU. In (f) and (g), we remove tracking entirely to assess its impact. Both show a substantial performance drop, especially in mIoU, indicating that tracking is essential. (g) includes momentum, while (f) does not. In (h), we test forward-only tracking instead of bidirectional. It performs worse than Ours (-0.029 in $S_\alpha$ and -0.053 in mIoU), showing bidirectional tracking is more effective. Finally, we remove mean subtraction from Ours, as shown in (i), which resulted in a slight increase in $S_\alpha$ and mIoU but a lower $E_\phi$. This means that subtraction has a minimum impact on performance. This might be due to the limited number of videos in the dataset featuring camera motion with relatively static objects, or because the pipeline relies more effectively on contrast rather than absolute color for object identification. Compared to other methods, our approach without mean subtraction (i) achieved the highest $S_\alpha$ and mIoU scores. However, our full method obtained the highest $E_\phi$, with $S_\alpha$ and mIoU scores close to those of (i).

Our final model includes all components: optical flow, background subtraction, mean subtraction, momentum, and bidirectional tracking. It achieves strong performance with $S_\alpha = 0.776$, $E_\phi = 0.878$, and mIoU = 0.550.

\subsection{Prompting Experements}
In supplementary material Section~\ref{sec:prompt}, we studied the effects of different Owlv2 prompts and SAM-2 prompts. Results show that when given prior knowledge and the color of the highlight to Owlv2, the detection performs the best, and when given boxes and points as prompts to SAM-2, the segmentation performs the best. In that section, we also argued why using prior knowledge is a fair comparison with previous methods.

\subsection{Data Contamination Concerns}
When using large foundational models, data contamination is a valid concern. However, we examined the training data timeline of Owlv2 and concluded that the contamination from the MoCA dataset is highly limited. A detailed explanation can be found in the supplementary material Section \ref{sec:contamination}.

%% file: sec/5_con.tex
\section{Conclusion}
We introduced ZS-VCOS, a zero-shot method for video camouflaged object segmentation, integrating optical flow, vision-language models, and SAM. Our approach significantly outperformed existing methods, increasing mIoU on the MoCA-Mask dataset from 0.273 to 0.561 and improving detection success on MoCA-Filtered from 0.628 to 0.697. Our findings highlight the potential of zero-shot pipelines for effectively handling camouflaged objects, particularly beneficial in scenarios lacking labelled data. Our modular design enables easy replacement of improved modules at any pipeline stage, enhancing overall performance.

Our method has several limitations. First, it is designed for videos containing one and only one object. In multi-object scenarios, the tracking and matching components would require modification to handle multiple object associations. Second, the approach relies on a textual description of the target object. While this is significantly less costly than collecting annotated training data, generating an accurate and unambiguous prompt can still be non-trivial in some cases. Potential solutions include incorporating few-shot object detection using example image embeddings from VLM as queries, or integrating an image-to-text captioning tool to automatically generate prompts from reference frames.



\section*{Acknowledgement}
This work was supported by NFRF GR024801 and CFI GR024473. We also thank Weathon Software (\url{https://weasoft.com}) for providing computing credits via Google Colab.

%% file: sec/X_suppl.tex
\clearpage
\setcounter{page}{1}
\maketitlesupplementary
\subsection{Data Contamination Concerns}
\label{sec:contamination}
When using large foundational models, data contamination is a valid concern. However, OWLv2 is trained using pseudo-labels from WebLI \cite{webli} generated by OWL-ViT \cite{owlvit}. This is less concerning, as OWL is primarily trained on image-text pairs without localization information. The detection data used in OWL-Vit was sourced from Object365 \cite{Shao_Li_Zhang_Peng_Yu_Zhang_Li_Sun_2019}, Open Images \cite{Benenson_Ferrari_2022}, and Visual Genome \cite{Krishna_Zhu_Groth_Johnson_Hata_Kravitz_Chen_Kalantidis_Li_Shamma_eta}, all of which were published before the original MoCA \cite{Lamdouar_Yang_Xie_Zisserman_2020} dataset. Therefore, while contamination is a consideration, it is highly limited. At the time of this paper's publication, two new datasets \cite{tavu2024camovid, Gao_Feng_Wang_Hong_Zhou_Fei_Wang_Zhang_2025} (not yet released) could provide uncontaminated data, and we encourage future work to evaluate our method on these datasets. 

Data contamination is an acknowledged issue when using large-scale foundational models. However, in our case, while it is possible that visual content from MoCA may have appeared in pretraining corpora, it is highly unlikely that the segmentation ground truth or specific frame-level annotations were included. Therefore, even under \textbf{worst-case} assumptions, the problem reduces to a transductive inference setting. Our pipeline remains zero-shot in the sense that no ground truth of the target task is used in any training stage.

\section{Previous Metrics Inconsistance}
\label{res:metric}
Metrics such as mean IoU can be computed in three primary ways: (1) calculating IoU for each frame individually, averaging across frames within one video, and then averaging across videos, (2) calculating IoU for each frame, and averaging all frames' results, or (3) calculating an IoU for all frames, which is equivalent to treating the entire video sequence as a single, large concatenated image for both the predicted masks and the ground truth masks, and then computing the IoU on these two large, combined frames. The three calculation methods can yield different results, occasionally significant. Additionally, the calculation script for each paper varies slightly, such as SLT-Net omits the last frame to keep for flow-based methods. 
 In our paper, we computed the metrics using the SLT-Net script. This methodology might differ from other reported methods. We recalculated these metrics using the SLT-Net evaluation implementation for the supervised state-of-the-art method, ZoomNeXt, and it resulted in slightly elevated results of $E_m$=0.755, mDice=0.511, and mIoU=0.438 compared to the original reporting (see Table~\ref{tab:res}). However $S_a$, $F_{\beta}^w$, and MAE remains unchanged. Thus, in this paper, we focus on these 3 metrics when comparing them across previous methods.

\begin{table}
    \centering
    \begin{tabular}{c|ccc}
    \hline
         Video&  ZoomNeXt& SLTNet&Ours
\\
         \hline
         (arctic\_fox)&  0.812& 0.667&\textbf{0.842}\\
         (arctic\_fox\_3)&  0.347& 0.251&\textbf{0.787}\\
         (black\_cat\_1)&  0.429& 0.31&\textbf{0.479}\\
         (copperhead\_snake)&  0.061& 0.359&\textbf{0.575}\\
         (flower\_crab\_spider\_0)&  \textbf{0.881}& 0.112&0.761
\\
         (flower\_crab\_spider\_1)&  \textbf{0.835}& 0.643&0.783
\\
         (flower\_crab\_spider\_2)&  \textbf{0.812}& 0.605&0.758
\\
         (hedgehog\_3)&  \textbf{0.55}& 0.288&0.502
\\
         (ibex)&  0.271& 0.168&\textbf{0.615}\\
         (mongoose)&  \textbf{0.413}& 0.314&0.388
\\
         (moth)&  0.519& 0.534&\textbf{0.774}\\
         (pygmy\_seahorse\_0)&  0.064& \textbf{0.149}&0.0
\\
         (rusty\_spotted\_cat\_0)&  0.233& \textbf{0.269}&0.217
\\
         (sand\_cat\_0)&  \textbf{0.772}& 0.281&0.613
\\
         (snow\_leopard\_10)&  0.001& 0.001&\textbf{0.468}\\
         (stick\_insect\_1)&  0.004& 0.003&\textbf{0.246}\\
         \hline
    \end{tabular}
    \caption{mIoU of each video in the MoCA-Mask test set.}
    \label{tab:video}
\end{table}

\begin{table*}
    \centering
    \begin{tabular}{c|>{\centering\arraybackslash}p{20mm}>{\centering\arraybackslash}p{20mm}>{\centering\arraybackslash}p{20mm}|ccc}
    \hline
         &  Mention of ``animal or insect"&Mention Of Highlight&Negative Prompts&  $S_{\alpha}$&  $E_{\phi}$& mIoU\\
         \hline
         (a)&    &\Checkmark&\Checkmark&  0.570&  0.623& 0.205\\
 (b)& \Checkmark& & \Checkmark& 0.749& 0.868&0.501\\
         (c)&  \Checkmark&\Checkmark&&  \textbf{0.776}&  \textbf{0.878}& \textbf{0.550}\\
    
 (d)& \Checkmark& \Checkmark& \Checkmark& \textbf{0.776}& \textbf{0.878}&\textbf{0.550}\\
      \hline
    \end{tabular}
    \caption{Effects of different VLM prompting strategies}
    \label{tab:prompt}
\end{table*}

\begin{table}
    \centering
    \begin{tabular}{c|c|ccc}
    \hline
         &  SAM-2 Prompt&  $S_{\alpha}$&  $E_{\phi}$& mIoU\\
         \hline
         (a)&  Box Only&  \textbf{0.776}&  0.873& 0.540\\
         (b)&  Point + Box&  \textbf{0.776}&  \textbf{0.878}& \textbf{0.550}\\
         \hline
    \end{tabular}
    \caption{Effects of different SAM-2 prompts}
    \label{tab:my_label}
\end{table}
\subsection{Prompting Experements}
\label{sec:prompt}
To investigate different prompting strategies for the VLM, we tested performance (1) without explicitly asking the model to look for animals or insects, (2) without asking it to look for highlights, and (3) without negative prompts. Same as in previous sections, we swept the VLM threshold from 0.03 to 0.13 with steps of 0.02.

\subsubsection{Prior Knowledge in Prompt}
In our methods, we explicitly instructed the vision-language model to look for animals or insects highlighted in blue. However, to evaluate the model's generalization capability, we replaced the specific terms ``animal or insect" in the prompts with the more generic term "object." The corresponding results are shown in Table~\ref{tab:prompt} row (a). As observed, performance drops significantly when switching from specific categories (animal/insect) to a general object category. We suspect this is because multiple objects are often moving within the video, making it unclear to the model which objects it should focus on. 

We do not consider the use of prompts mentioning animals to disqualify our method as zero-shot. Camouflaged videos often contain multiple moving or camouflaged elements—such as leaves, lighting, or branches—making it ambiguous for a model to determine which object should be segmented without explicit guidance. The model cannot ``mind-read" our purpose of the current test. For example, if we are now looking for non-animal objects in the image, the model has no way of knowing this information. Providing a very general and high-level prior (e.g., ``animal or insect") is essential for disambiguating the target in the absence of supervision. Previous work claiming zero-shot, like \cite{Zhou_Sun_Li_Benini_Konukoglu_2024}, also used a similar prior in their prompt (``Please provide the coordinates of the bounding box where the animal is camouflaged in the picture"). Previous work that has been trained on MoCA-Mask could effectively learn this information from the dataset, and previous work that has not been trained on MoCA-Mask (Like FlowSAM \cite{flowsam} and Motion Grouping \cite{Yang_Lamdouar_Lu_Zisserman_Xie_2021}) has been trained to identify the center, large moving object. These methods introduced the prior knowledge by training, making it fair to compare against our method with prompt prior knowledge. Moreover, these models with fixed prior knowledge might be harder to transfer to other domains (e.g., non-animal objects or videos without a center and big objects) without finetuning. We chose the animal dataset MoCA because it is currently the only large-scale, publicly available dataset for video camouflage segmentation. Other datasets that include non-animal camouflaged objects, such as MSVCOD \cite{gao_msvcod}, have not been released at the time of this work. Future work testing the generalizability of these methods and our methods is needed. 

\subsubsection{Mentioning Highlight Color in Prompt}
To emphasize motion within the frame, we highlighted the relevant areas in blue. Our ablation study demonstrated the effectiveness of the highlighting itself. Additionally, we explicitly guided the visual language model (VLM) by prompting it to detect animals or insects highlighted in blue. The impact of this prompt was tested by removing it, as shown in row (c) of Table~\ref{tab:prompt}, where the prompt was simply "an animal or insect." Without explicitly instructing the model to focus on highlighted areas (indicative of motion), we observed a slight performance drop across all metrics. Nonetheless, performance remained relatively high, suggesting that highlighting motion regions alone aids detection, even without explicit prompting (see comparison with row (b), no highlighting and no instruction for highlighting, in Table~\ref{tab:ab}). 

\subsubsection{Negative Prompt}
We hypothesized that negative prompts could help the model avoid negative objects. However, as shown in Table~\ref{tab:prompt} row (c), the results are identical to the setting with negative prompts. (Note that although these two settings achieved the same best performances, their results are not identical at all VLM threshold settings.) This shows that the VLM used (Owlv2 \cite{Minderer_Gritsenko_Houlsby_2024}) can effectively avoid non-targeted interest without negative prompting.